\documentclass[letterpaper, 10 pt, conference]{ieeeconf}
\IEEEoverridecommandlockouts
\usepackage{amsmath,amssymb,amsfonts}
\usepackage{algpseudocode}
\usepackage{algorithm}
\usepackage{algpseudocode}
\usepackage{graphicx}
\usepackage{textcomp}
\usepackage[usenames, dvipsnames, table]{xcolor}
\usepackage{makecell}
\usepackage{multirow}
\usepackage{overpic}
\usepackage{pict2e}
\usepackage[utf8]{inputenc}
\usepackage[style=numeric-comp,sorting=none,backend=bibtex]{biblatex}
\def\BibTeX{{\rm B\kern-.05em{\sc i\kern-.025em b}\kern-.08em
    T\kern-.1667em\lower.7ex\hbox{E}\kern-.125emX}}
    
\usepackage{hyperref}
\usepackage[capitalize]{cleveref}
\usepackage{amssymb}
\usepackage{multirow}
\usepackage{amssymb}
\usepackage{pifont}
\usepackage[mathscr]{euscript}
\usepackage{subcaption}
\usepackage[normalem]{ulem}
\useunder{\uline}{\ul}{}
\newcommand{\cmark}{\ding{51}}%
\newcommand{\xmark}{\ding{55}}%

\usepackage{tikz}

\newcommand\copyrighttext{
  \footnotesize \textcopyright 2023 IEEE. Personal use of this material is permitted.
  Permission from IEEE must be obtained for all other uses, in any current or future
  media, including reprinting/republishing this material for advertising or promotional
  purposes, creating new collective works, for resale or redistribution to servers or
  lists, or reuse of any copyrighted component of this work in other works.}
\newcommand\copyrightnotice{%
\begin{tikzpicture}[remember picture,overlay]
\node[anchor=south,yshift=10pt] at (current page.south) {\fbox{\parbox{\dimexpr\textwidth-\fboxsep-\fboxrule\relax}{\copyrighttext}}};
\end{tikzpicture}%
}

\begin{document}

\title{CAT-RRT: Motion Planning that Admits Contact\\ One Link at a Time}

\author{Nataliya Nechyporenko$^{*}$, Caleb Escobedo, Shreyas Kadekodi, and Alessandro Roncone %
\thanks{*Authors are with the Human Interaction and Robotics [HIRO] Group, Computer Science Department, University of Colorado Boulder, Boulder, CO USA. This work is partially supported by NSF FW-HTF grant \#2222952/2953. NN is supported by NSF DGE \#2040434. {\tt\small name.surname@colorado.edu}}
\thanks{This paper was published at the IEEE/RSJ International Conference on Intelligent Robots and Systems (IROS) 2023. 
DOI: \href{https://ieeexplore.ieee.org/document/10341668}{0.1109/IROS47612.2022.9982198}}
}

\maketitle
\thispagestyle{empty}
\pagestyle{empty}
\copyrightnotice
\addtolength{\belowcaptionskip}{-2.5pt}
\begin{abstract}
Current motion planning approaches rely on binary collision checking to evaluate the validity of a state and thereby dictate where the robot is allowed to move. This approach leaves little room for robots to engage in contact with an object, as is often necessary when operating in densely cluttered spaces. In this work, we propose an alternative method that considers contact states as high-cost states that the robot should avoid but can traverse if necessary to complete a task.
More specifically, we introduce \textsl{Contact Admissible Transition-based Rapidly exploring Random Trees} (CAT-RRT)\footnote{Supplementary video and open source code \cite{projectlink}.}, a planner that uses a novel per-link cost heuristic to find a path by traversing high-cost obstacle regions. Through extensive testing, we find that state-of-the-art optimization planners tend to over-explore low-cost states, which leads to slow and inefficient convergence to contact regions. Conversely, CAT-RRT searches both low and high-cost regions simultaneously with an adaptive thresholding mechanism carried out at each robot link. This leads to paths with a balance between efficiency, path length, and contact cost.
\end{abstract}
\section{Introduction} \label{introduction}
\begin{figure}[t]
    \begin{overpic}[width=0.48\linewidth,percent]
    {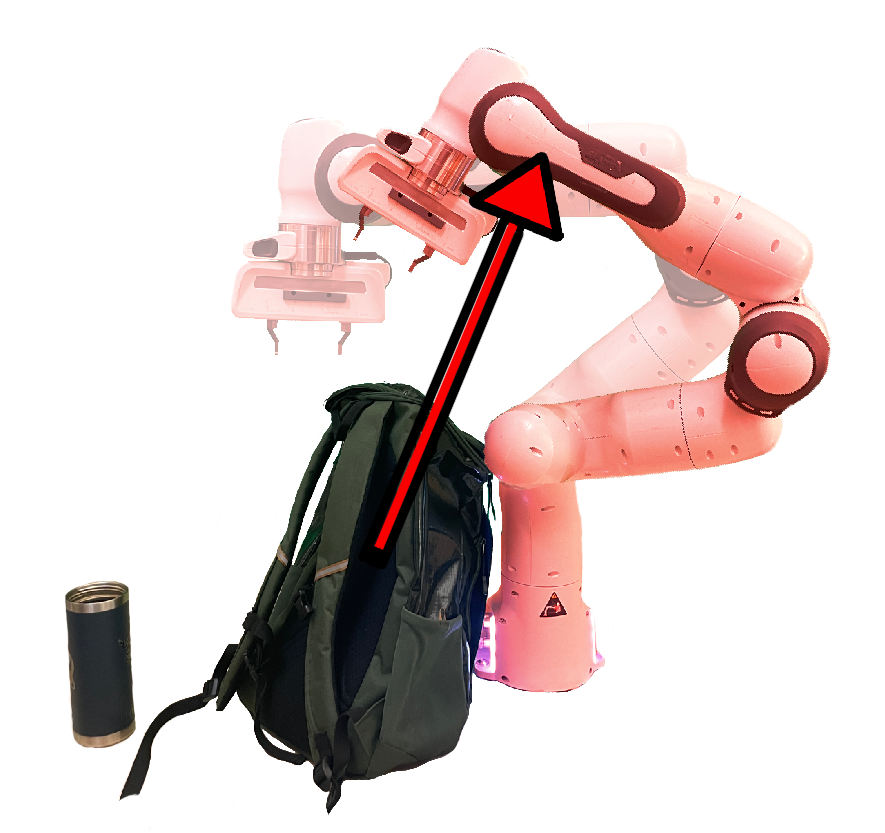}%
    \put(5, 32){\textcolor{ForestGreen}{Goal}}
    \put(52, 4){\color{red}Obstacle}
    \put(30, 92){\color{black}Whole Arm Cost}
    \put(124, 92){\color{black}Per-link Cost}
    \put(90, -2)
    {\includegraphics[width=0.51\linewidth]{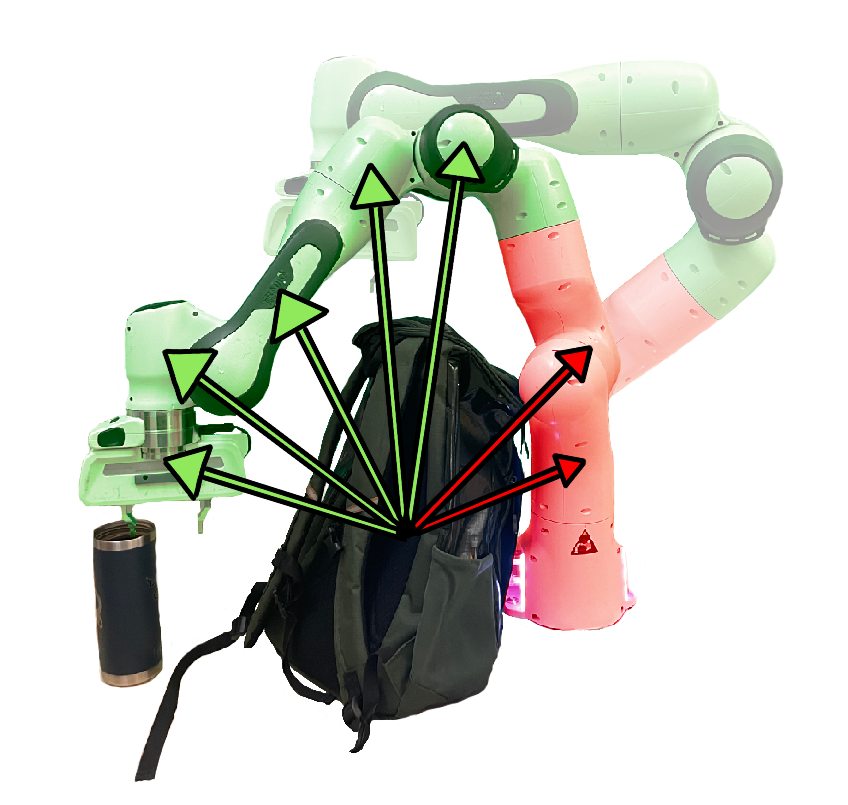}}
    \end{overpic}\vspace{-4pt}
    \caption{CAT-RRT is an optimization planner which uses a per-link cost heuristic to generate a path in clutter by 
    allowing contact to occur if it is necessary to succeed at the task.
    Rather than invalidating contact states or restricting motion for the entire arm (left), we propose a method that generates a path by prioritizing the least impacted links (right).}
    \label{fig:first-fig}
    \vspace{-15pt}
\end{figure}
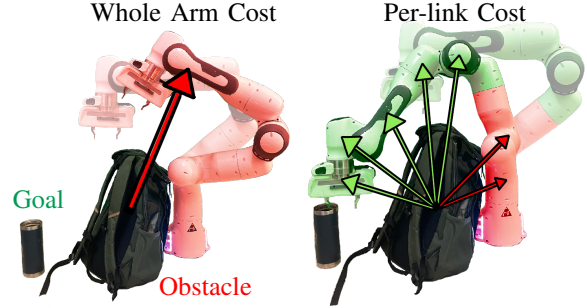

Robot behaviors are designed around the fundamental safety constraint of collision-free paths, as it ensures minimal physical interaction with the environment that could lead to robot error states or damage. 
However, this principle is oftentimes too limiting, as environmental constraints (e.g. tight spaces, areas with occlusion), perceptual constraints (e.g. narrow field of view, sensor inaccuracies), and operational constraints (e.g. maintaining a vertical cup orientation to avoid spilling) must also be accounted for while guaranteeing a collision-free path. 
As a result, a robot manipulator will likely fail to reach into a cluttered space due to the minimal clearance between the arm and the objects required to meet collision-free guarantees (see \cref{fig:first-fig}).
Because motion planning is a fundamental component of a robot operating in the real world, having it restricted means significantly hindering robot capabilities; this limits the potential for robots to complete real-world tasks in unstructured or semi-structured environments such as harvesting fruit on a farm or picking items in a cluttered warehouse. 

In this work, we are motivated by the idea that a binary collision test with a measure of whether the robot is in collision with the environment is insufficient to delineate the boundary between a valid or an invalid motion plan.
Collision checkers provide the motion planner with a query function to test whether two geometric models overlap \cite{pan2012fcl, jimenez2005collision}. Rather than invalidating any interactions between the robot and the environment, it is possible to evaluate them based on a continuous scale of object contact. This allows a robot to consider paths that would be discarded by traditional motion planning techniques while increasing success rate and enabling the robot to explore the environment through contact. 

More specifically, in this paper we introduce a motion planner, Contact Admissible Transition-based Rapidly exploring Random Trees (\textit{CAT-RRT}), that can generate paths in cluttered and unstructured environments by guiding the robot through states of \textsl{admissible contact}, which we define as contact necessary to reach the goal configuration. We are inspired by the literature in optimization-based motion planning \cite{amato1998choosing,wein2008planning}, which differs from traditional search-based motion planning in that it seeks to find a path that optimizes over a cost function. In particular, Transition-based Rapidly Exploring Random Tree (T-RRT, \cite{jaillet2008transition, jaillet2010sampling}) uses the output of a cost function to increment or decrement a single global variable, called \textsl{temperature}, which is proportional to the likelihood of accepting high-cost states. CAT-RRT differs from T-RRT by not only using a set of temperatures, but also having each branch within the search tree adjust their own temperatures. This allows the planner to simultaneously propagate paths into low and high-cost regions based on multiple variables. We define cost with respect to proximity or contact with an obstacle, as shown in \cref{fig:first-fig}.

\newcommand\boxdim{0.18}
\newcommand\ytop{54}
\newcommand\ybot{-48}
\newcommand\xone{120}
\newcommand\xtwo{222}
\newcommand\xthree{324}
\newcommand\xfour{426}

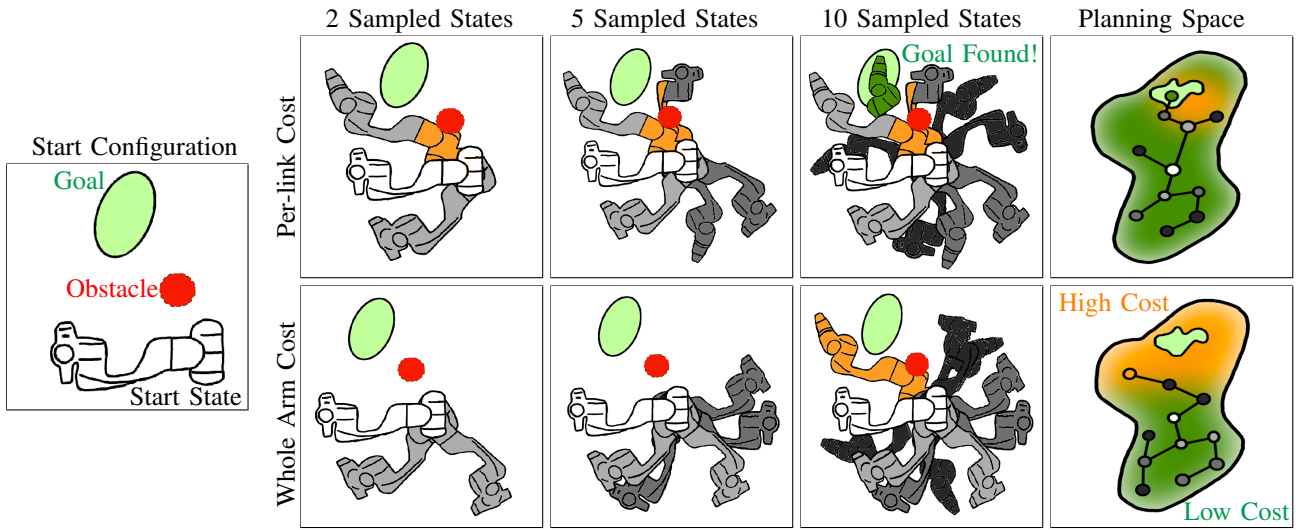
\begin{figure*}[t!]
    \vspace{+65pt}
  \frame{\begin{overpic}[width=\boxdim\linewidth,percent]
    {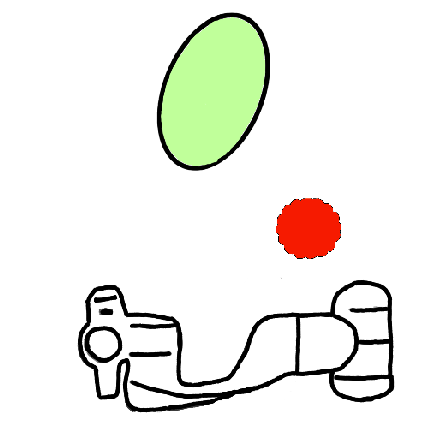}%
        \put(19, 90){\textcolor{ForestGreen}{Goal}}
        \put(24, 46){\color{red}Obstacle}
        \put(50, 2){\color{black}Start State}
        \put(10, 104){\color{black}Start Configuration}
        \put(130, 156){\color{black}2 Sampled States}
        \put(230, 156){\color{black}5 Sampled States}
        \put(332, 156){\color{black}10 Sampled States}
        \put(437, 156){\color{black}Planning Space}
        \put(110, 70){\rotatebox{90}{\color{black}Per-link Cost}}
        \put(110, -40){\rotatebox{90}{\color{black}Whole Arm Cost}}
        
        \put(\xone,\ytop)
        {\frame{\includegraphics[width=\boxdim\linewidth,height=\boxdim\linewidth]{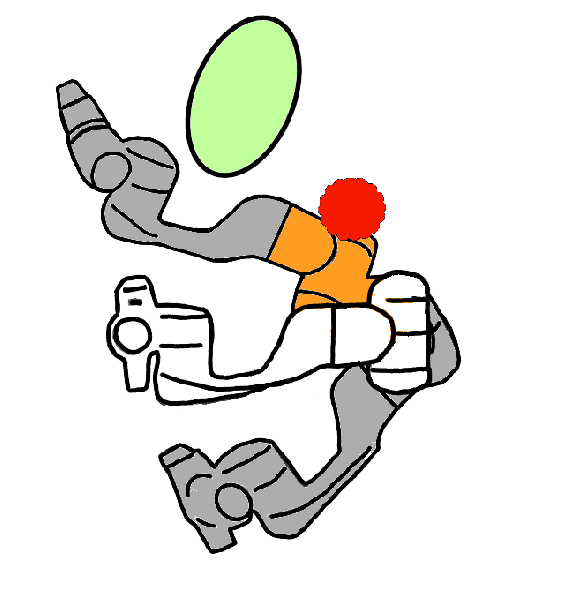}}}
        \put(\xone,\ybot)
        {\frame{\includegraphics[width=\boxdim\linewidth,height=\boxdim\linewidth]{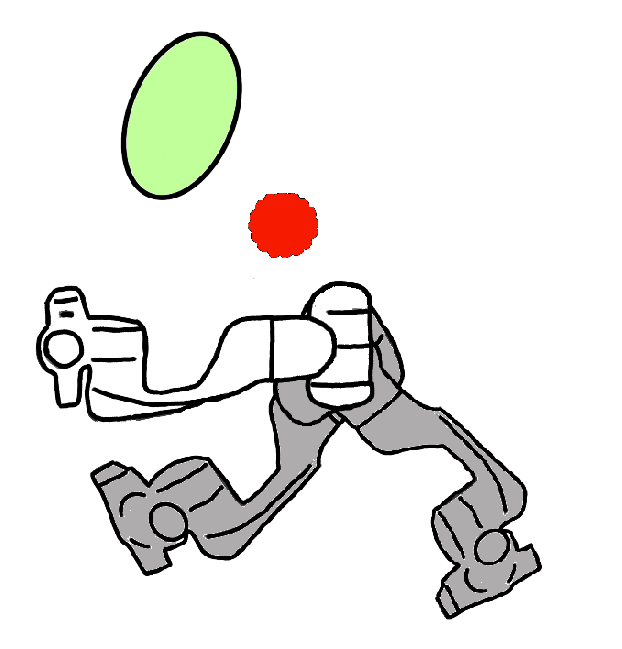}}}
        \put(\xtwo,\ytop)
        {\frame{\includegraphics[width=\boxdim\linewidth,height=\boxdim\linewidth]{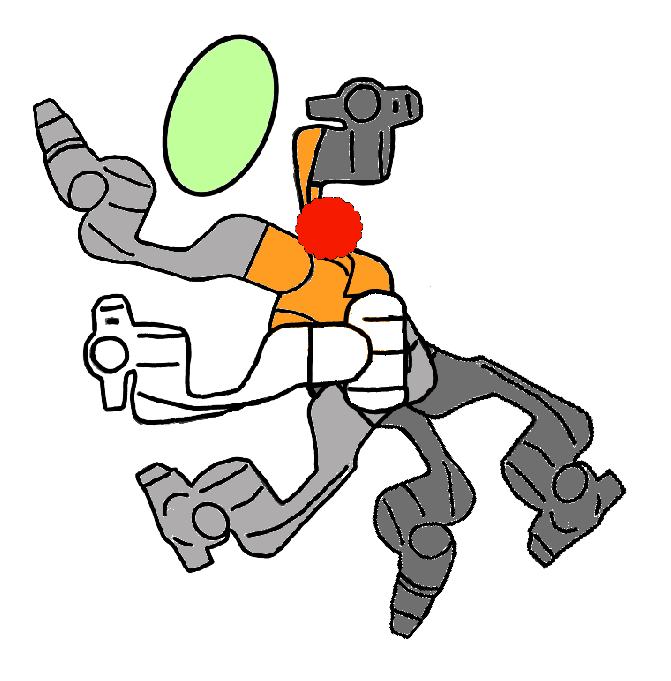}}}
        \put(\xtwo,\ybot)
        {\frame{\includegraphics[width=\boxdim\linewidth,height=\boxdim\linewidth]{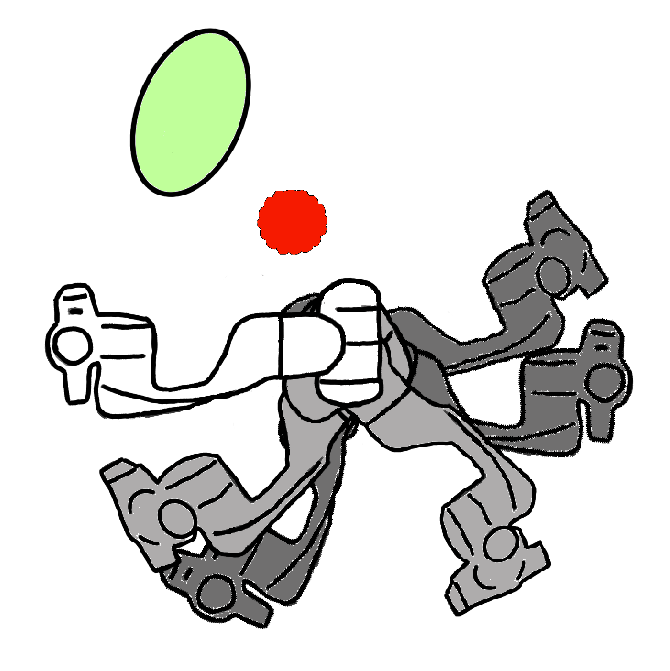}}}
        \put(\xthree,\ytop)
        {\frame{\includegraphics[width=\boxdim\linewidth,height=\boxdim\linewidth]{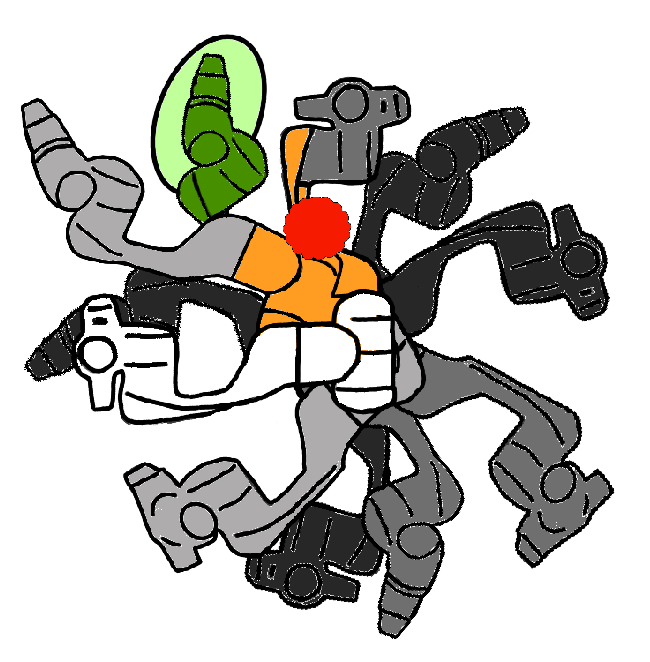}}
        }
        \put(365, 142){\textcolor{ForestGreen}{Goal Found!}}
        \put(\xthree,\ybot)
        {\frame{\includegraphics[width=\boxdim\linewidth,height=\boxdim\linewidth]{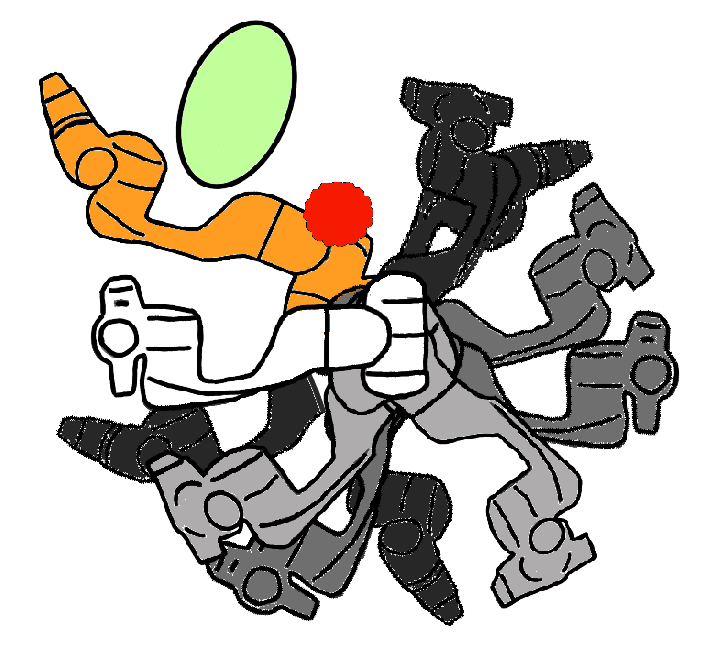}}}
        \put(\xfour,\ytop)
        {\frame{\includegraphics[width=\boxdim\linewidth,height=\boxdim\linewidth]{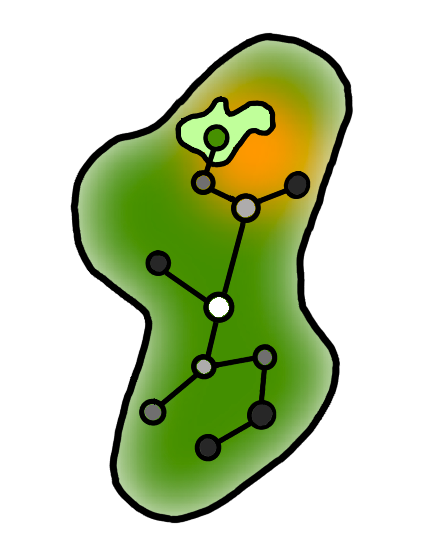}}}
        \put(\xfour,\ybot)
        {\frame{\includegraphics[width=\boxdim\linewidth,height=\boxdim\linewidth]{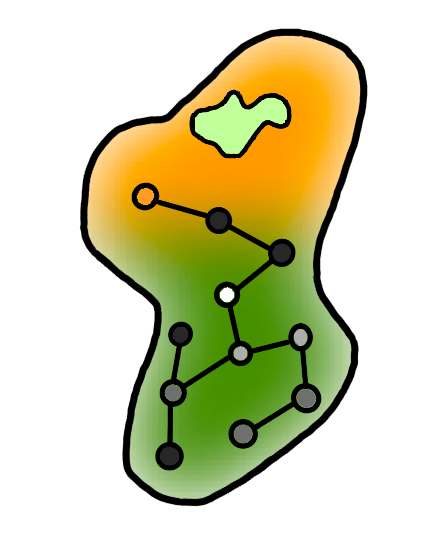}}}
        \put(429, 40){\color{orange}{High Cost}}
        \put(480, -45){\textcolor{ForestGreen}{Low Cost}}
    \end{overpic}}
    \vspace{+50pt}
    \caption{Example scenario for \textsl{per-link} (top row) and \textsl{whole-arm} cost (bottom row) with a common start configuration (leftmost vignette). Object--robot contact is shown in orange and sampled states are grouped in gray scale depending on when the state was sampled, darker states are sampled later in time. The rightmost column depicts the planning space of both cost heuristics with green being a low cost area and orange being high cost. The per-link cost planner is able to find the goal location due to a reduced high cost area surrounding the goal location even though some links are in contact with the object.}
    \label{fig:panda}
\end{figure*}

Additionally, in prior work cost functions are often defined to minimize travel distance \cite{amato1998choosing}, as short paths are a desired property \cite{wein2008planning}. In our work, we define a novel per-link cost heuristic which computes artificial repulsive and attractive vectors, together forming an artificial potential field (APF) \cite{khatib1986potential}, for every arm link. This allows the planner to assign a different temperature to every link and prioritize motion of the links that are least impacted by vector repulsion, as shown in \cref{fig:panda}.

In summary, our contributions are:
1) a novel optimization planner which successfully generates feasible trajectories even when the robot may need to come in contact with an obstacle;
2) an APF-based per-link cost heuristic which prioritizes motion with links that are unrestricted by contact. 
We performed an extensive quantitative evaluation in simulation and a qualitative demonstration in the real world. Collectively, our results demonstrate that, while relevant literature struggles to generate any path into high-cost regions, CAT-RRT can consistently find feasible trajectories by gradually admitting contact one link at a time.
\section{Related Work} \label{related-work}
In this section, 
we analyze three major approaches branching from optimization-based Rapidly Exploring Random Tree (RRT): informed approaches, stochastic approaches, and node-changing approaches. We give a brief overview of the representative planners we choose from each category to use as baselines for our work. We also summarize several works that explore contact admissible motion planning without a focus on optimization.

In the wake of success of sampling-based planners, RRT has been widely adopted due to its simplicity and efficiency \cite{elbanhawi2014sampling, lavalle1998rapidly, lavalle2001randomized}. However, because any feasible path is accepted without regard for path quality, it generally produces sub-optimal solutions \cite{gammell2021asymptotically}.
To improve upon RRT, other works propose a method of prioritizing nodes that converge toward an optimal solution \cite{gammell2021asymptotically, karaman2011sampling}. Often, this is achieved with an informed heuristic during node creation or a modified acceptance test that uses path quality to bias nodes toward low cost regions. 
The most prominent of these is RRT*, which is used as one of the baselines in our evaluation. RRT* is an incremental sampling-based planning algorithm that maintains a tree without any ``redundant'' edges---edges that are not within the lowest cost path from the start to current node in the tree \cite{karaman2011sampling}. 
RRT*, like other tree refinement methods, has optimality guarantees.
There are several existing modifications of RRT* as well, including using potential field-guided RRT* sampling, but these have not been tested on high-dimensional robot systems \cite{noreen2018optimal, wang2020improved, mohammed2021rrt}. 

A new wave of batch-informed trees have been proposed, which focus on both efficiency and path quality \cite{ holston2017fast, gammell2020batch, strub2020advanced}. 
One such planner is Batch-Informed Tree* (BIT*), used in our evaluation, which leverages a local optimization module to improve an initial path toward a local optimum.
BIT* is probabilistically complete and has been shown to find solutions more often than other almost-surely asymptotically optimal planners. 
Other optimization methods include stochastic planning algorithms. 
One such example is Transition-based RRT (T-RRT), which propagates a tree search based on a stochastic optimization method with transition tests to accept or reject new states, but offers no optimality guarantees \cite{jaillet2010sampling, jaillet2008transition}. 
Other works build on T-RRT to enable anytime behavior, bi-directional tree growth, and applicability to multi-agent systems 
\cite{devaurs2013enhancing, devaurs2014multi, devaurs2015optimal, iehl2012costmap}.
CAT-RRT shares a T-RRT-like optimization approach but with a unique transition test (detailed in Section \ref{sec:catrrt}). To demonstrate this distinction, we use T-RRT as a baseline in our evaluation.

Several planners attempt to improve optimization efficiency by biasing sampled nodes based on a chosen direction \cite{berenson2011addressing, nemlekar2021robotic}. This strategy is desirable because the search can be moved in the direction of low-cost regions especially when guided by potential fields. 
Vector Field RRT (VF-RRT) does this through the Upstream Criterion, as defined in \cref{eq:UpstreamCriterion}, which is used to bias sampling toward nodes that minimize the extent to which a path goes against a given vector field \cite{ko2014randomized}. We chose VF-RRT to evaluate this strategy since it is highly applicable to potential field-based cost functions which our work relies on.

The following two papers are the closest to our work. \cite{nemlekar2021robotic} develops a potential field guided RRT* algorithm for the problem of fruit harvesting \cite{nemlekar2021robotic}. It defines leaves as permeable obstacles, which the robot is allowed to come into contact with after incurring a cost. The authors use a combination of an RRT*-like approach with tree refinement and a VF-RRT-like approach with node biasing---both of which are evaluated in our experimental framework.
Finally, \cite{kabutan2018motion} compares a potential field cost function as applied to T-RRT and other sampling approaches \cite{kabutan2018motion}. The authors do not consider contact behaviors and rely on a simplified cost calculation between a single point on the robotic arm and an arbitrary obstacle point. 

Finally, several works address contact admissibility and motion planning in the context of perception. Instead of using optimization, these works replace a binary collision-check function with a binary cost-based function \cite{bhattacharjee2014robotic, park2014interleaving, saund2020motion}. They rely on a threshold that reflects how much contact a robot can make with an object. Such a threshold is challenging to define ahead of time for all environments. In contrast, our work uses an adaptive threshold mechanism.

\section{Methods} \label{methods}
In this section, we first outline the problem of path finding. Next, we describe how CAT-RRT plans a path while optimizing over a cost function using a set of temperatures and a transition test. Then, we describe how the temperatures are generated based on a separate cost for each link of the arm. Finally, we define additional cost heuristics from existing literature, which are used to evaluate CAT-RRT.

\subsection{Problem description}
We use a similar definition of the planning problem as \cite{gammell2020batch}. Let $Q \subseteq \mathbb{R}^n$ be the state space of the planning problem. Let $\mathbf{q}_{\text {start }} \in Q_{\text {free }}$ be the initial state of joint angles and $\mathbf{q}_{\text {goal }} \subset Q_{\text {free }}$ be the set of the desired goal states. Let $\sigma:[0,1] \rightarrow Q_{\text {free }}$ be a continuous map to a sequence of states through a space of bounded variation that can be executed by the robot (i.e. self-collision free, feasible path) and $\Sigma$ be the set of all such nontrivial paths. The optimal planning problem is then formally defined as the search for a path, $\sigma^* \in \Sigma$, that minimizes a given cost function, $c: \Sigma \rightarrow \mathbb{R}^{n}_{\geq 0}$, while connecting  $\mathbf{q_{\text{start}}}$ to $\mathbf{q}_{\text {goal }} \in Q_{\text {goal }}$ where $\mathbb{R}^{n}_{\geq 0}$ is the set of non-negative real numbers. 

\subsection{Motion planning with CAT-RRT} \label{sec:catrrt}
CAT-RRT benefits from the exploratory strength of RRT-like algorithms that quickly expand toward large regions of unexplored space. Additionally, it integrates features of stochastic optimization methods from T-RRT-like planners, which use transition tests to accept or reject potential states. The main algorithm runs as follows: a random state, $\mathbf{q}_{rand}$, is selected from the configuration space, which is a minimum distance away from $\mathbf{q}_{near}$. A transition test function is used to evaluate $\mathbf{q}_{rand}$. If it passes the transition test, then it is added to the tree, and the process repeats until a path to $\mathbf{q}_{goal}$ is found. The main tree construction algorithm of CAT-RRT is defined in \cite{jaillet2008transition} and will not be reintroduced here for brevity. However, the transition test is unique to our approach and defined in \cref{alg:TTA}. First, we evaluate a vector of costs, $\mathbf{C}$, for $\mathbf{q}_{rand}$, with each cost corresponding to a link on the arm. Next, we obtain a vector of temperatures, $\mathbf{T}$, stored in the nearest node of the tree. One link at a time, we evaluate and update the tree node based on a transition test. A transition test is passed if the link's cost, $\mathbf{C}[i]$, is lower than its allowed temperature, $\mathbf{T}[i]$, and the temperature is reduced unless it reaches a user-defined minimum value, $t_{min}$. If all the links pass the test, then the temperature vector is stored in the child node of the added state. A failed transition test increases the temperature for the given link, thereby increasing the chance of a state sampled in that region to be accepted in the next iteration. 
Although previous works use an intermediate exponential function based on the Metropolis criterion to relate cost and temperature \cite{jaillet2008transition}, we did not find this beneficial and opted for a direct relationship. In our algorithm, the temperature is synonymous to a dynamic cost threshold. Both $\omega$ and $\gamma$ are user-defined values that control the rate of temperature decrement and increment. Our source code provides more specifics on parameter tuning \cite{projectlink}.

\begin{algorithm}
\caption{CAT-RRT Transition Test}\label{alg:TTA}
\begin{algorithmic}
\State $\mathbf{C} \gets $\normalfont{\fontfamily{qcr}\selectfont GetPerLinkCost}$(\mathbf{q}_{near}, \mathbf{q}_{rand}, \mathbf{q}_{goal})$ \Comment{Eq. 4}
\State $\mathbf{T} \gets $\normalfont{\fontfamily{qcr}\selectfont GetTemperature}$(Node_{parent})$
\For{$i=0 ... L$}
\If{$\mathbf{C}[i] < \mathbf{T}[i]$ and $\mathbf{T}[i] > t_{min}$}
    \State $\mathbf{T}[i] -= \omega$ 
\ElsIf{$\mathbf{C}[i] > \mathbf{T}[i]$}
    \State $\mathbf{T}[i] += \gamma$
    \State \textbf{return} $False$
\EndIf
\EndFor
\State \normalfont{\fontfamily{qcr}\selectfont StoreTemperature}$(\mathbf{T}, Node_{child})$
\State \textbf{return} $True$
\end{algorithmic}
\end{algorithm}

CAT-RRT differs from T-RRT in that, rather than having a global temperature parameter for all nodes, the temperature is stored at the parent node and inherited by the child node. Furthermore, rather than storing a temperature as a scalar for the entire robot body, we create a temperature vector where each link is independently represented. Consequently, the tree accepts or rejects nodes based on the temperature at every link. This results in CAT-RRT's two distinct properties: 1) each branch of the tree regulates its own temperature, and 2) each link on the robotic arm enters high cost regions independent from the rest of the kinematic chain. When one link is in a high-cost region, it will stay in this position while the other links maintain low-cost region positioning. In the real world, this equates to one link of the robotic arm maintaining contact with an object while the other links continue to traverse contact-free spaces. This is in contrast to planners that attempt to always maintain low costs throughout the arm which may lead to scattered contact along a path. \cref{fig:panda} summarizes how CAT-RRT converges to a goal state using discrete costs along the robotic arm. In the absence of obstacles, CAT-RRT's transition test is not invoked and the planner operates as RRT. Next, we describe how the robot's perception of the environment is converted to a cost for each link. 

\begin{figure}[t]
    \begin{overpic}[width=0.45\linewidth,percent]
    {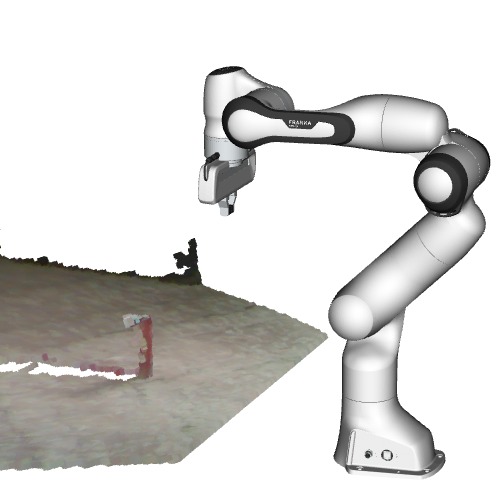}%
    \put(120, -2)
    {\includegraphics[width=0.46\linewidth]{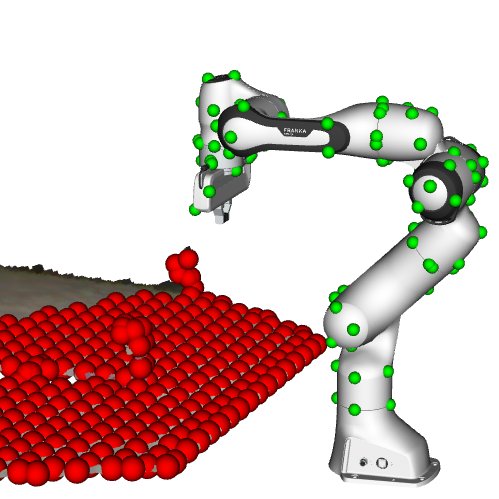}
    }
    \put(165, 85){\color{black}{Control Points}}
    \put(110, 50){\color{black}{Point Obstacles}}
    \put(197, 80){\linethickness{0.4mm}\color{black}\line(-10, 10){5}}
    \put(149, 40){\linethickness{0.4mm}\color{black}\line(-10, 10){9}}
    \end{overpic}
    \caption{The image on the left shows the robot in front of a point cloud of an object sitting on top of a table. The image on the right shows the same scene with an overlay of point obstacles in red and robot control points in green.}
    \label{fig:plc}
    \vspace{-4pt}
\end{figure}
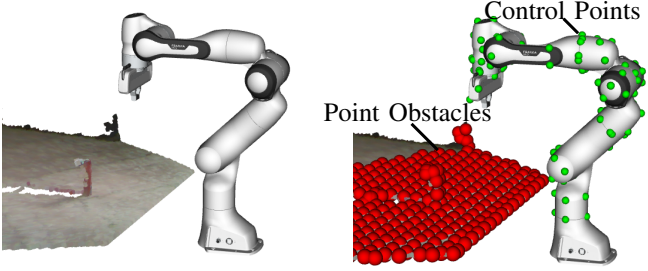

In this work, we try to step away from the dependence on high-resolution collision models for motion planning, as these are often unavailable for a robot operating in unstructured settings. However, each planner does require a basic understanding of the environment and the robot's position in space. To acquire this understanding, as detailed in \cref{fig:plc}, we assume the robot is equipped with a camera that relays depth perception information, as a point cloud, to the motion planning algorithm. The point cloud is converted to \textsl{point obstacles}, which are used as the basis for the planner's obstacle representation. \cref{fig:plc} shows the original point cloud and point obstacles, $p_k \in \Lambda$. Similarly, to ease reliance on 3D mesh models, the planner uses a set of \textsl{control points} to represent the robot and its position. The control points, $p_q \in \Gamma$, are represented by green spheres on \cref{fig:plc}. 

\subsection{Defining the cost heuristics}
\subsubsection{Controlling cost magnitude}
Repulsive vector costs and unit magnitude costs serve as the basis of the cost heuristics defined in this paper. These costs are generated from the distance between point obstacles and robot control points. The vector magnitudes, $\vec{\mathbf{v}}$, are scaled to be inversely proportional to the distance. Rather than opting for the traditional potential field equation introduced by Khatib et al. \cite{khatib1986potential}, which increases the repulsive force to infinity as the distance to obstacles becomes zero, we use a scaling function, $\boldsymbol{S}$, shown in \cref{eq:dist_scale}. Both $a$ and $b$ are scalar parameters, which allow us to control the magnitude of cost associated with contact. Here, $a$ controls the maximum scaling value of $\vec{\mathbf{v}}$ and $b$ controls how fast the function converges to the maximum as $||\vec{\mathbf{v}}||$ goes to zero. 
\begin{equation}\label{eq:dist_scale} \small
    \boldsymbol{S(\vec{\mathbf{v}})} = \frac{a*\vec{\mathbf{v}}}{b*||\vec{\mathbf{v}}||+1}
\end{equation}

\begin{figure}[t]
\vspace{10pt}
\centering
\includegraphics[width=0.3\textwidth]{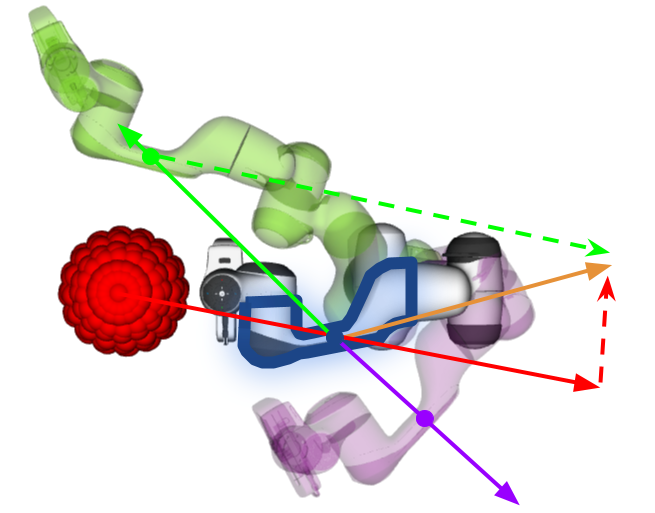}
    \put(-8, 55){\large\color{black}$\vec{\mathbf{v}}_{l_6}$}
    \put(-40, 10){\large\color{black}$\vec{\mathbf{d}}_{l_6}$}
    \put(-140, 77){\large\color{black}$\mathbf{q}_{goal}$}
    \put(-78, 4){\large\color{black}$\mathbf{q}_{rand}$}
    \put(-150, 50){\large\color{black}$\Lambda$}
    \put(-105, 31){\large\color{black}$\mathbf{q}_{near}$}
\caption{Robot's initial configuration ($\mathbf{q}_{near}$) in white, goal configuration ($\mathbf{q}_{goal}$) in green, random sampled state ($\mathbf{q}_{rand}$) in purple, and a set of point obstacles ($\Lambda$) in red. The directional vector for link number six, $\vec{\mathbf{d}}_{l_6}$, points from $\mathbf{q}_{near}$ to $\mathbf{q}_{rand}$. The desired directional vector for the link,  $\vec{\mathbf{v}}_{l_6}$, is a weighted sum between the vector from $\Lambda$ to $\mathbf{q}_{near}$ and the vector from $\mathbf{q}_{near}$ to $\mathbf{q}_{goal}$.}
\label{fig:contact-vf}
\end{figure}
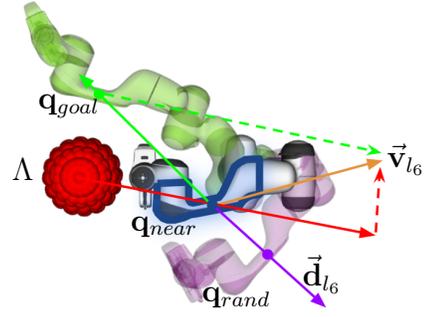

\subsubsection{Per-link cost heuristic used with CAT-RRT} 
The per-link cost heuristic is an essential component of CAT-RRT as it guides the search tree.
The desired vector at link \textit{l}, $\vec{\mathbf{v}}$, defines the desired direction of motion in Cartesian space for every link of the arm. $K$ is the number of point obstacles, $N$ is the number of control points, $L$ is the number of links, $p_k$ is the $k$th obstacle point, $p_{q_{near},i}$ is the $i$th control point on the robot's $\mathbf{q}_{near}$ state, and link number $l$ is $l \in [1, ..., L]$, $p_{q_{goal},i}$ is the \textit{i}th control point on the robot's goal state. $\alpha$ and $\beta$ are scalar parameters. 
\begin{multline}\label{eq:Vvf} \small
    \vec{\mathbf{v}} = \frac{1}{K} \sum_{k=1}^{K} \frac{1}{N} \sum_{i=1}^{N}  \alpha *\mathbf{S}(p_{q_{near},i} - p_k)\\ + \beta*(p_{q_{goal},i} - p_{q_{near},i})
\end{multline}
The random directional vector $\vec{\mathbf{d}}$ from $\mathbf{q}_{near}$ to the uniformly sampled state $\mathbf{q}_{rand}$ at every link is obtained as follows;
\begin{equation}\label{eq:Dvf} \small
    \vec{\mathbf{d}}= \frac{1}{N} \sum_{i=1}^{N} (p_{q_{rand},i}-p_{q_{near},i})
\end{equation}
The cost at every link $\mathbf{c}$ is defined by \cref{eq:Coverlap}, with lower costs indicating an alignment between the directional vector and the desired vector.
\begin{equation}\label{eq:Coverlap} \small
    \mathbf{c} =  (-\vec{\mathbf{v}}) \cdot \vec{\mathbf{d}}
\end{equation}

\cref{fig:contact-vf} shows the components of the per-link vector field alignment cost heuristic, which guides the robot away from obstacles and towards goal locations using randomly sampled states. Next, we define the cost heuristics from previous work and how we implement them. These methods are used by state-of-the-art planners for comparison against CAT-RRT and the per-link cost heuristic.

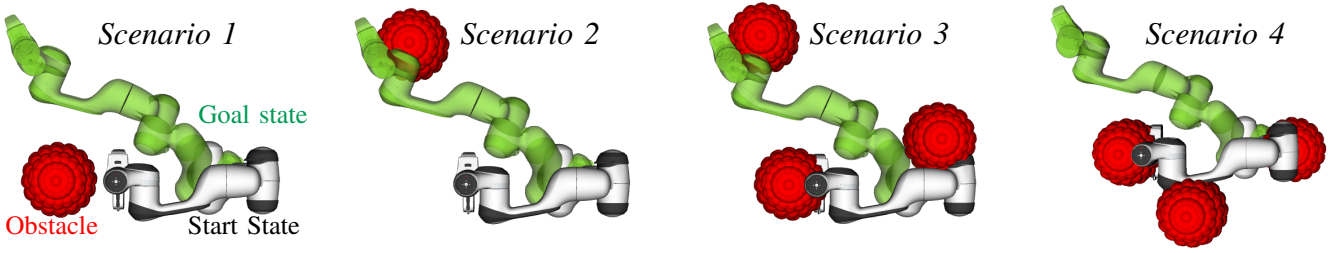
\begin{figure*}[t]
    \vspace{10pt}
    \centering
        \begin{minipage}{.21\textwidth}
            \begin{subfigure}{\textwidth}
            \centering
            \includegraphics[width=\textwidth]{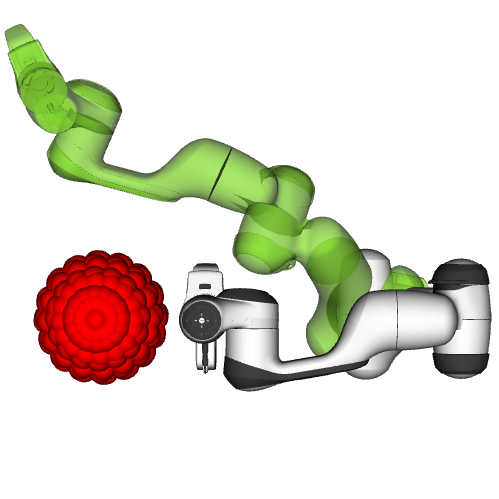}
            \put(-70, 90){\large\color{black}\textit{Scenario 1}}
            \put(-36, 18){\color{black}Start State}
            \put(-105, 18){\color{red}Obstacle}
            \put(-32, 60){\textcolor{ForestGreen}{Goal state}}
            \end{subfigure}\\
        \end{minipage}
        \hfill
        \begin{minipage}{.21\textwidth}
            \begin{subfigure}{\textwidth}
            \centering
            \includegraphics[width=\textwidth]{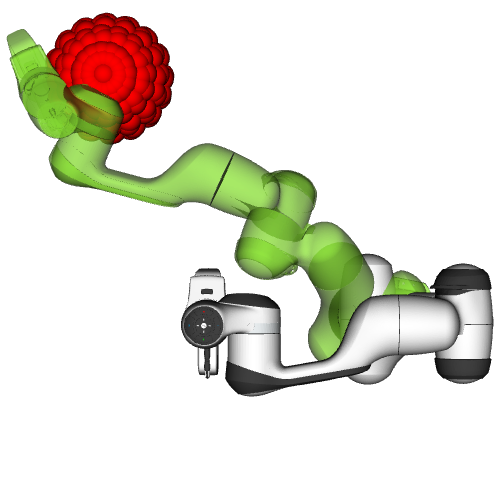}
            \put(-65, 90){\large\color{black}\textit{Scenario 2}}
            \end{subfigure}\\
        \end{minipage}
        \hfill
        \begin{minipage}{.21\textwidth}
            \begin{subfigure}{\textwidth}
            \centering
            \includegraphics[width=\textwidth]{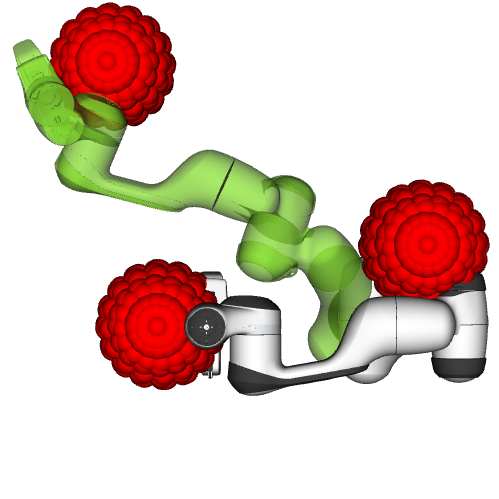}
            \put(-65, 90){\large\color{black}\textit{Scenario 3}}
            \end{subfigure}\\
        \end{minipage}
        \hfill
        \begin{minipage}{.21\textwidth}
            \begin{subfigure}{\textwidth}
            \centering
            \includegraphics[width=\textwidth]{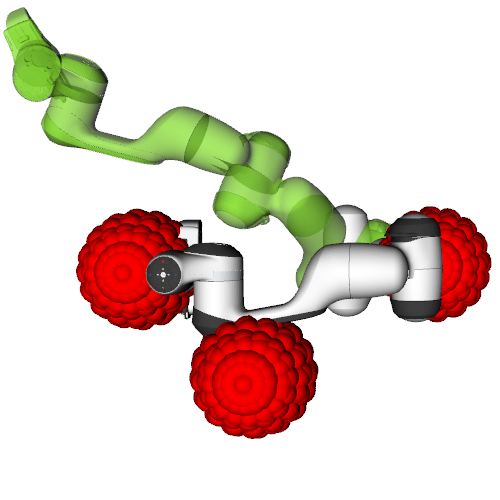}
            \put(-70, 90){\large\color{black}\textit{Scenario 4}}
            \end{subfigure}\\
        \end{minipage}
        \vspace{-28pt}
        \caption{In our evaluation, the robot is tasked with finding a path from the start state (white) to the goal state (green) while moving through obstacle regions (red) in four experimental scenarios of increasing complexity. The scenarios from left to right are increasingly more complex with obstacles overlapping with the start and goal states.}
        \label{fig:four-scenarios}
\end{figure*}

\subsection{Comparison with state of the art}
\subsubsection{Obstacle overlap heuristic used with T-RRT, RRT*, and BIT*}\label{sec:overlap-opt-method} The ``permissible contact" planners \cite{bhattacharjee2014robotic, park2014interleaving, saund2020motion} detailed in \cref{related-work} evaluate the cost of a path based on the amount of overlap between the robot and the potential obstacles in the environment. 
In this work, the amount of obstacle--robot overlap is equivalent to adding up the vector magnitudes given from \cref{eq:dist_scale}, which is implemented as the cost function $\mathbf{C}$:
\newcommand\norm[1]{\left\lVert#1\right\rVert}
\begin{equation}\label{eq:Cvf} \small
    \mathbf{C} =  \sum_{l=1}^{L} \norm{\frac{1}{K} \sum_{k=1}^{K} \frac{1}{N} \sum_{i=1}^{N}\mathbf{S}(p_k - p_{q_{near},i})}
\end{equation}
 This baseline cost heuristic is used to generate low-cost paths by T-RRT, RRT*, and BIT*. 

\subsubsection{Upstream Criterion used with VF-RRT} As discussed in Section \ref{related-work}, one approach to improve the convergence rate of sampling-based planners is to adjust the newly sampled nodes in the direction of a vector field \cite{nemlekar2021robotic, berenson2011addressing, ko2014randomized}. To test this approach, we use VF-RRT with the Upstream Criterion \cite{ko2014randomized}. 
The Upstream Criterion is defined as: \begin{equation}\label{eq:UpstreamCriterion}
    \int_{0}^{L}(\lvert\lvert f(q(s)) \rvert\rvert - \langle f(q(s)), q^{'}(s) \rangle) ds
\end{equation} 
where $f(q(s))$ is a piecewise continuous vector field and $\lvert\lvert f(q(s)) \rvert\rvert$ represents the norm of $\langle f(q(s)), q^{'}(s) \rangle$.
The function $f(q(s))$ is not explicitly defined in the original paper and is left for the user to define based on a specific application. Since we are planning in robot configuration space, the output of $f(q(s))$ must be a vector of joint angles. However, our robot and the environment are defined in Cartesian space by point obstacles and control points. To obtain a set of joint angles from a set of points in Cartesian space, we apply the inverse of the Jacobian $\mathbf{J}$ to \cref{eq:dist_scale}.
$\mathbf{J}_l^\dagger \in \mathbb{R}^{3xl} $ is the Moore-Penrose pseudo-inverse of $\mathbf{J}$ at link \textit{l}.
\begin{equation}\label{eq:vfrrt-cost} \small
   f(q(s)) =  \sum_{l=1}^{L} \left( \mathbf{J}_l^\dagger \times \left(  \frac{1}{K} \sum_{k=1}^{K} \frac{1}{N} \sum_{i=1}^{N} \left( \mathbf{S}(p_k - p_{q_{near},i}\right) \right) \right)
\end{equation}
\section{Experiments} \label{experiments}
The experimental evaluation is performed in both simulation (\cref{subsec:simulation}) and real-world (\cref{subsec:real-world}). 
The former allows for repeatable and reproducible experiments, while the latter shows the applicability of planning with contact in the real world. 

\subsection{Specifications of the experimental testbed}
Based on the discussion in \cref{related-work} and the implementation in \cref{methods}, we evaluate: T-RRT, RRT*, and BIT*, which use the obstacle-robot overlap cost heuristic, VF-RRT, which uses the upstream criterion, and CAT-RRT, which uses the per-link cost heuristic. Each planner was allotted a maximum of 60 seconds to compute and refine a path. We believe this to be a reasonable amount of evaluation time and comparable to prior work---e.g.  \cite{gammell2020batch} used a 20 second limit for a similar 7 degree-of-freedom (DOF) problem to evaluate BIT* with limited compute power. 

Each planner relies on a set of control points and point obstacles referred to in \cref{sec:catrrt}. To obtain the control points, we extract 115 vertices from the robot's 3D mesh, openly available on the Franka Emika repository. To obtain point obstacles, we downsample a point cloud using the Point Cloud Library Voxel Grid \cite{rusu20113d} filter with a leaf size of 0.05m. The point cloud is then converted to the robot's frame of reference. Each of the resulting voxels, or values on a regular grid in 3D space, is considered as a point obstacle. In simulation, the point obstacles are added artificially to create example objects, represented by red orbs on Scenarios 1-4 in \cref{fig:four-scenarios}.

The planning algorithms are implemented in C++ with the the ROS (Noetic) framework \cite{quigley2009ros}. We use TRRT, VF-RRT, RRT*, and BIT* within the Open Motion Planning Library \cite{sucan2012open} and integrate CAT-RRT within the library as well. We use Moveit! for simulation \cite{chitta2016moveit} and Rviz for visualization. Franka Emika Panda is used as the robot platform and an OAK-D Pro \cite{oakd} camera to capture the point cloud. All experiments were performed on a computer with an Intel i9 processor and 16GB RAM.

\subsection{Simulated experimental scenarios} \label{subsec:simulation}
For the simulated experiments, four scenarios of increasing complexity are designed---see \cref{fig:four-scenarios}. Scenario 1 evaluates if each planner can succeed in finding a path from a free low-cost start state to a free low-cost goal state in the presence of a single obstacle. This is a baseline scenario used to check fundamental path finding capabilities. In Scenario 2, the planner is asked to compute a path in which the robot's goal state is in contact with an obstacle. This scenario tests the planner's ability to plan into a high-cost region. Scenario 3 includes two obstacles at the start state and one in the goal state, which tests the planner's ability to traverse between two high-cost regions. Finally, Scenario 4 is set up similarly to Scenario 3, but with an additional obstacle blocking the path away from the other obstacles, meaning the robot cannot break contact with the high-cost regions as in Scenario 3. This tests how the planner is able to modulate high-cost regions across the robotic arm. 

\subsection{Metrics for evaluation}
We evaluate each planner based on its ability to successfully generate a path within the allotted time. For the resulting trajectories, we measure the distribution of contact along the arm and the overall path length. These metrics represent the planner's ability to minimize contact cost while moving toward the goal. We run fifty trials for each planner in each scenario and average the metrics across the successful trials. Path length is calculated as the sum of the $L^2$-norm between the end-effector Cartesian points of the trajectory. For the simulated experiments, we measure the amount of contact along the trajectory by calculating the overlap between the 3D mesh of the arm and the obstacles. This is done through a collision post-processing step. First, we place spherical collision objects of the same size as the red point obstacle orbs into the robot scenario. Then, we run collision detection based on the Bullet Physics Engine on every state along the trajectory. The collision checker returns the number of states in collision and the contact depth, or penetration depth, between each overlapping robot-obstacle pair.  
\section{Results and Discussion} \label{results}
In this section, we summarize the results obtained after running the experiments outlined in Section \ref{experiments}. We demonstrate that T-RRT and VF-RRT struggle to navigate into high-cost regions with contact. While RRT* tends to prioritize shorter paths by incurring more contact, BIT* generates longer paths with less contact. In contrast, CAT-RRT tends to find a better balance between path length and contact depth.  

\subsection{Simulation experiments}
\subsubsection{Scenarios 1 \& 2}
All planners are able to find a path with no obstacle overlap for Scenario 1. However, the results from Scenario 2 demonstrate a significant rift in the capabilities of the planners, in that T-RRT and VF-RRT were able to compute a successful path 0/50 times while the other planners were able to find such a path 50/50 times. Table \ref{T:scene12} summarizes the binary results of the planners in their ability to plan into high-cost regions. 

\begin{figure} [t]
    \centering
        \begin{minipage}{.475\columnwidth}
            \begin{subfigure}{\textwidth}
            \centering
            \includegraphics[width=\textwidth]{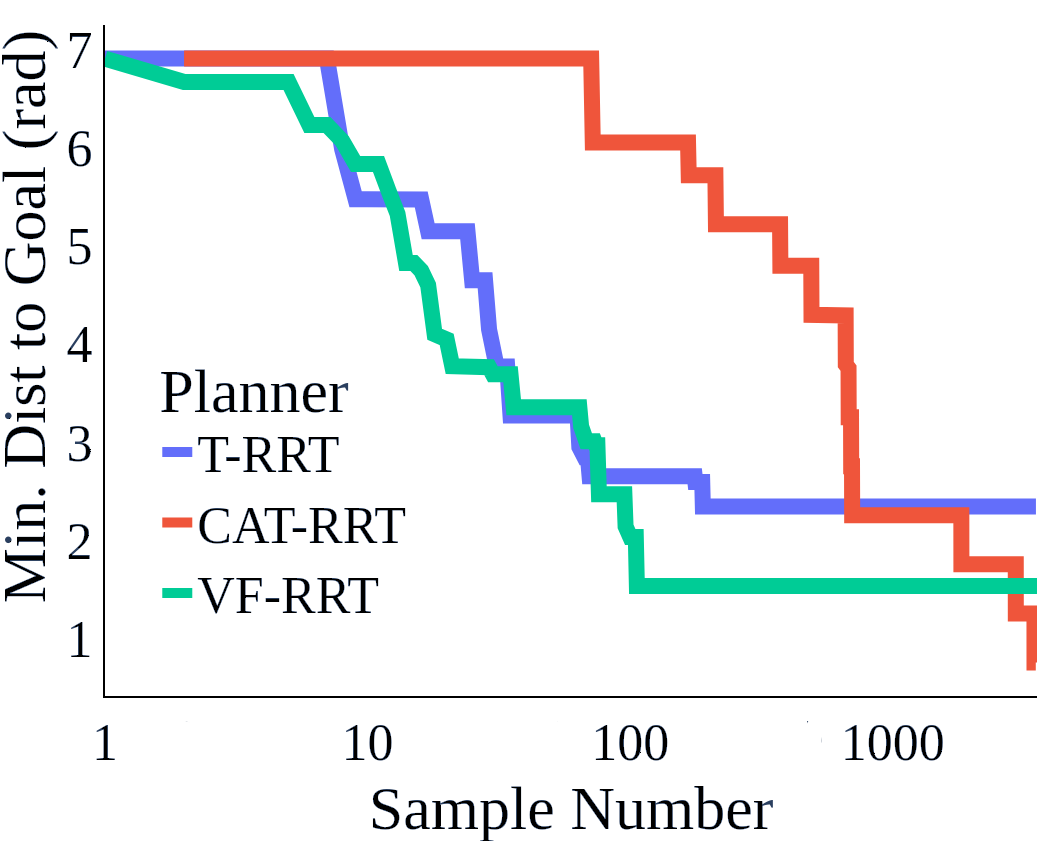}
            \end{subfigure}\\
        \end{minipage}
        \begin{minipage}{.475\columnwidth}
            \begin{subfigure}{\textwidth}
            \vspace{14pt}
            \centering
            \includegraphics[width=\textwidth]{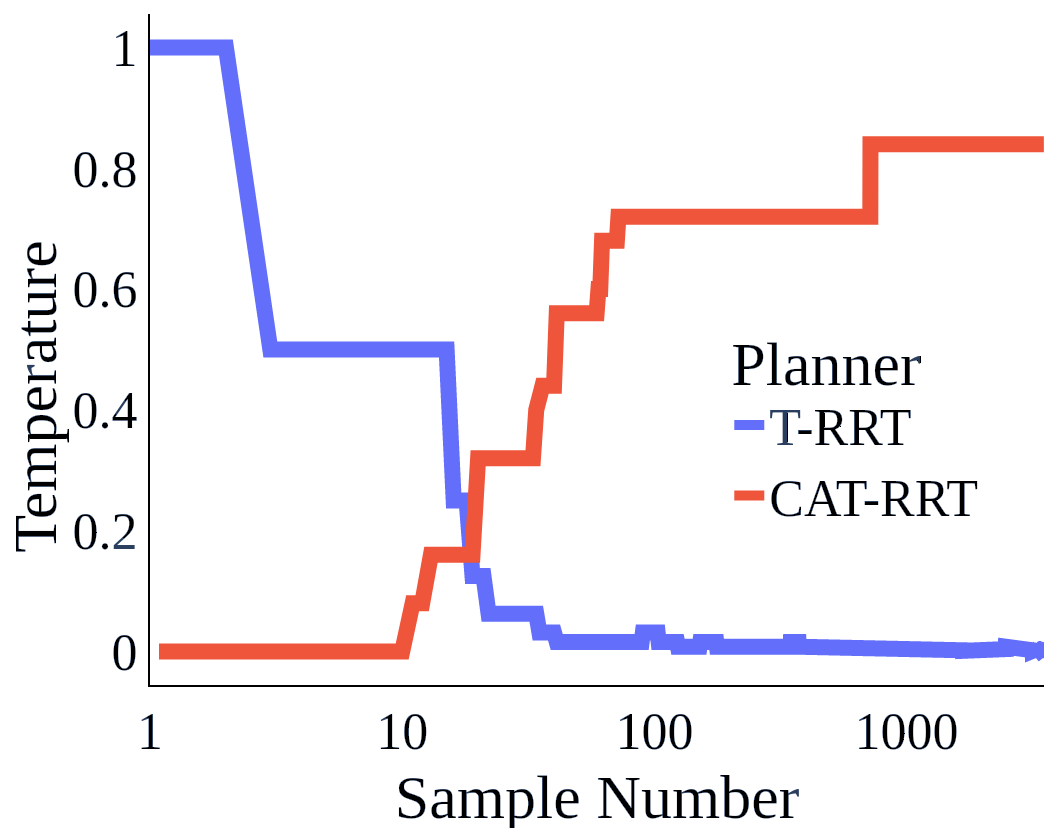}
            \label{fig:obstacle}
            \end{subfigure}\\
        \end{minipage}
        \vspace{-20pt}
        \caption{Scenario 2 analysis of T-RRT and VF-RRT failures. (Left) The minimum distance to goal of new nodes as T-RRT builds its tree. (Right) The average temperature and distance to goal of CAT-RRT. Unlike T-RRT and VF-RRT, CAT-RRT overcomes the high-cost threshold and converges to a solution.}
        \label{fig:trrt-temp-plot}
\end{figure}

The reason T-RRT struggles to find a path in Scenario 2 is because the global temperature variable is prohibitive in allowing the tree to explore high-cost regions. The temperature parameter is proportional to the probability of having a state accepted as a node in the tree. \cref{fig:trrt-temp-plot}, right shows that the temperature drops early in the iteration phase because of the large number of samples generated in the low-cost space of the robotic arm. This prevents the states in the high-cost regions near obstacles from being accepted. 

VF-RRT faces a similar issue: newly sampled states will always be directed away from the high-cost regions, which in Scenario 2 is where the goal state is located. Whereas a sample in a high-cost region of T-RRT would be rejected, with VF-RRT it would be adjusted downstream from the high-cost region and ultimately end up further from the goal state. Because T-RRT and VF-RRT cannot compute plans into high-cost regions within a reasonable amount of time given our testbed specification, we excluded them from our comparison chart in the subsequent tests in Scenario 3-4. In this sense, we use Scenario 2 to filter out methods which struggle to generate paths in complex cost spaces with high DOF robots. 

\begin{table}[]
\vspace{8pt}
\setlength{\tabcolsep}{2.5pt}
\renewcommand{\arraystretch}{1.5}
\begin{center}
\begin{tabular}{c|c|c|c|c|c|}
\cline{2-6}
 & \textbf{T-RRT} & \textbf{VF-RRT} & \textbf{RRT*} & \textbf{BIT*} & \textbf{CAT-RRT} \\ \hline
\multicolumn{1}{|c|}{\textbf{Scenario 1}} & \cmark & \cmark & \cmark & \cmark & \cmark \\ \hline
\multicolumn{1}{|c|}{\textbf{Scenario 2}} & \xmark & \xmark & \cmark & \cmark & \cmark \\ \hline
\end{tabular}
\end{center}
\caption{Here, \cmark represents the planner successfully finding a path 50/50 times in under 60 seconds and \xmark represents a failure or a 0/50 success rate of finding a path to the goal region.}
\label{T:scene12}
\end{table}

\begin{table*}[]
\vspace{8pt}
\begin{small}
\begin{center}
\setlength{\tabcolsep}{2.8pt}
\renewcommand{\arraystretch}{1.5}
\begin{tabular}{|c|ccccccccccc|ccccccccccc|}
\hline
 & \multicolumn{11}{c|}{\textbf{Scenario 3}} & \multicolumn{11}{c|}{\textbf{Scenario 4}} \\ \cline{2-23} 
 & \multicolumn{3}{c|}{\textbf{Path Metric}} & \multicolumn{8}{c|}{\textbf{Total Contact Depth for Link (mm)}} & \multicolumn{3}{c|}{\textbf{Path Metric}} & \multicolumn{8}{c|}{\textbf{Total Contact Depth for Link (mm)}} \\ \cline{2-23} 
\multirow{-3}{*}{\textbf{Method}} & \begin{tabular}[c]{@{}c@{}}S\\ (50)\end{tabular} & \begin{tabular}[c]{@{}c@{}}T\\ (s)\end{tabular} & \multicolumn{1}{c|}{\begin{tabular}[c]{@{}c@{}}PL\\ (m)\end{tabular}} & 1 & 2 & 3 & 4 & 5 & 6 & 7 & 8 & \begin{tabular}[c]{@{}c@{}}S\\ (50)\end{tabular} & \begin{tabular}[c]{@{}c@{}}T\\ (s)\end{tabular} & \multicolumn{1}{c|}{\begin{tabular}[c]{@{}c@{}}PL\\ (m)\end{tabular}} & 1 & 2 & 3 & 4 & 5 & 6 & 7 & 8 \\ \hline
\textbf{RRT*} & 36 & 62.9 & \multicolumn{1}{c|}{1.7} & 0. & 0. & 0.1 & 15.4 & 16.1 & 12.8 & 23.9 & 28.4 & 31 & 63.2 & \multicolumn{1}{c|}{1.1} & 0. & 18.1 & \cellcolor[HTML]{34CDF9}41.5 & 6.3 & \cellcolor[HTML]{F56B00}28.9 & 22. & \cellcolor[HTML]{9932CC}32.6 & \cellcolor[HTML]{34FF34}35.8 \\ \hline
\textbf{BIT*} & 50 & 60.0 & \multicolumn{1}{c|}{3.9} & 0. & 0. & 0.2 & 20.6 & 3.2 & 8.1 & 18.3 & 21.5 & 50 & 60.0 & \multicolumn{1}{c|}{1.8} & 0. & 23.2 & \cellcolor[HTML]{34CDF9}43.8 & 10.7 & \cellcolor[HTML]{F56B00}30.1 & 29.7 & \cellcolor[HTML]{9932CC}28.7 & \cellcolor[HTML]{34FF34}24.5 \\ \hline
\textbf{CAT-RRT} & 50 & 18.7 & \multicolumn{1}{c|}{1.2} & 0. & 0. & 0. & 18.3 & 0.2 & 5.5 & 13.9 & 25.4 & 50 & 15.9 & \multicolumn{1}{c|}{1.2} & 0. & 24.6 & \cellcolor[HTML]{34CDF9}51.2 & 5.6 & \cellcolor[HTML]{F56B00}10.2 & 13. & \cellcolor[HTML]{9932CC}20.8 & \cellcolor[HTML]{34FF34}22.9 \\ \hline
\end{tabular}
\end{center}
\vspace{-4pt}
\caption{Experimental results of each planning algorithm for Scenarios 3 and 4. The path metrics include the number of successes out of 50 trials (S), the average time to compute the path within the allotted time budget of 60s (T), and the total path length of the end-effector (PL). All the metrics, including contact depth for each link, are averaged across the successful trials. The highlighted colors in Scenario 4 correspond to the maximum contact depth peaks in Figure \ref{fig:scene4-results}.}
\label{T:scene34-results}
\end{small}
\end{table*}

\subsubsection{Scenario 3}
\begin{figure} 
    \centering
        \begin{minipage}{.32\columnwidth}
            \begin{subfigure}{\textwidth}
            \centering
            \includegraphics[width=\textwidth]{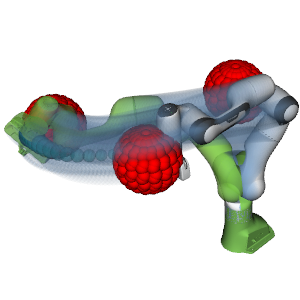}
            \put(-50, 75){\large\color{black}RRT*}
            \end{subfigure}\\
        \end{minipage}
        \begin{minipage}{.32\columnwidth}
            \begin{subfigure}{\textwidth}
            \centering
            \includegraphics[width=\textwidth]{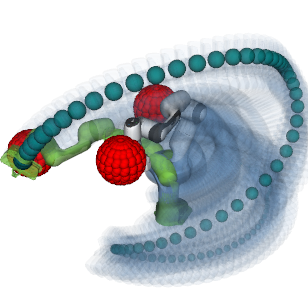}
            \put(-50, 75){\large\color{black}BIT*}
            \end{subfigure}\\
        \end{minipage}
        \begin{minipage}{.32\columnwidth}
            \begin{subfigure}{\textwidth}
            \centering
            \includegraphics[width=\textwidth]{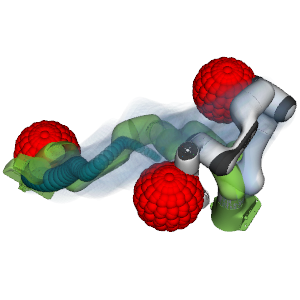}
            \put(-60, 75){\large\color{black}CAT-RRT}
            \end{subfigure}\\
        \end{minipage}   
        \vspace{-15pt}
        \caption{Example trajectories for Scenario 3 generated by each planner. RRT* chooses to traverse the obstacle in front, BIT* first moves away from all obstacles before returning in the direction of the goal state, and CAT-RRT finds a low-cost path in between the two obstacles.}
        \label{fig:scene3-contact}
\end{figure}

Table \ref{T:scene34-results} details simulation results from Scenario 3. BIT* finds a solution path that is double the length of RRT* and almost triple the length of CAT-RRT. However, BIT* does maintain lowest minimum contact across the links. RRT* has a shorter path length than BIT*, but at the expense of high contact depth. CAT-RRT tends to generate a short path length without moving into the obstacles region as much as RRT*. CAT-RRT also produces its paths much faster computationally. \cref{fig:scene3-contact} shows a sample end-effector trajectory taken by each planner in Scenario 3. RRT* chooses to move through the obstacle, BIT* takes a longer path but with low contact impact, and CAT-RRT chooses to navigate between the obstacles while optimizing for the cost at every link. Although we do not apply smoothing to any of the generated trajectories, CAT-RRT tends to suffer the most from this as it does not go through any rewiring steps as the other planners. A post-processing trajectory optimization step can smooth out the trajectory and reduce the higher average contact depth values for CAT-RRT in Scenario 3.

\subsubsection{Scenario 4}
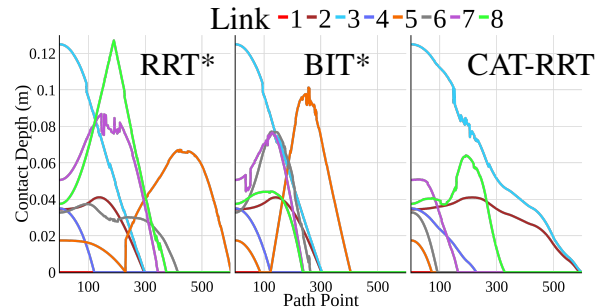
\begin{figure} 
    \centering
    \begin{overpic}[width=0.9\linewidth,percent]
    {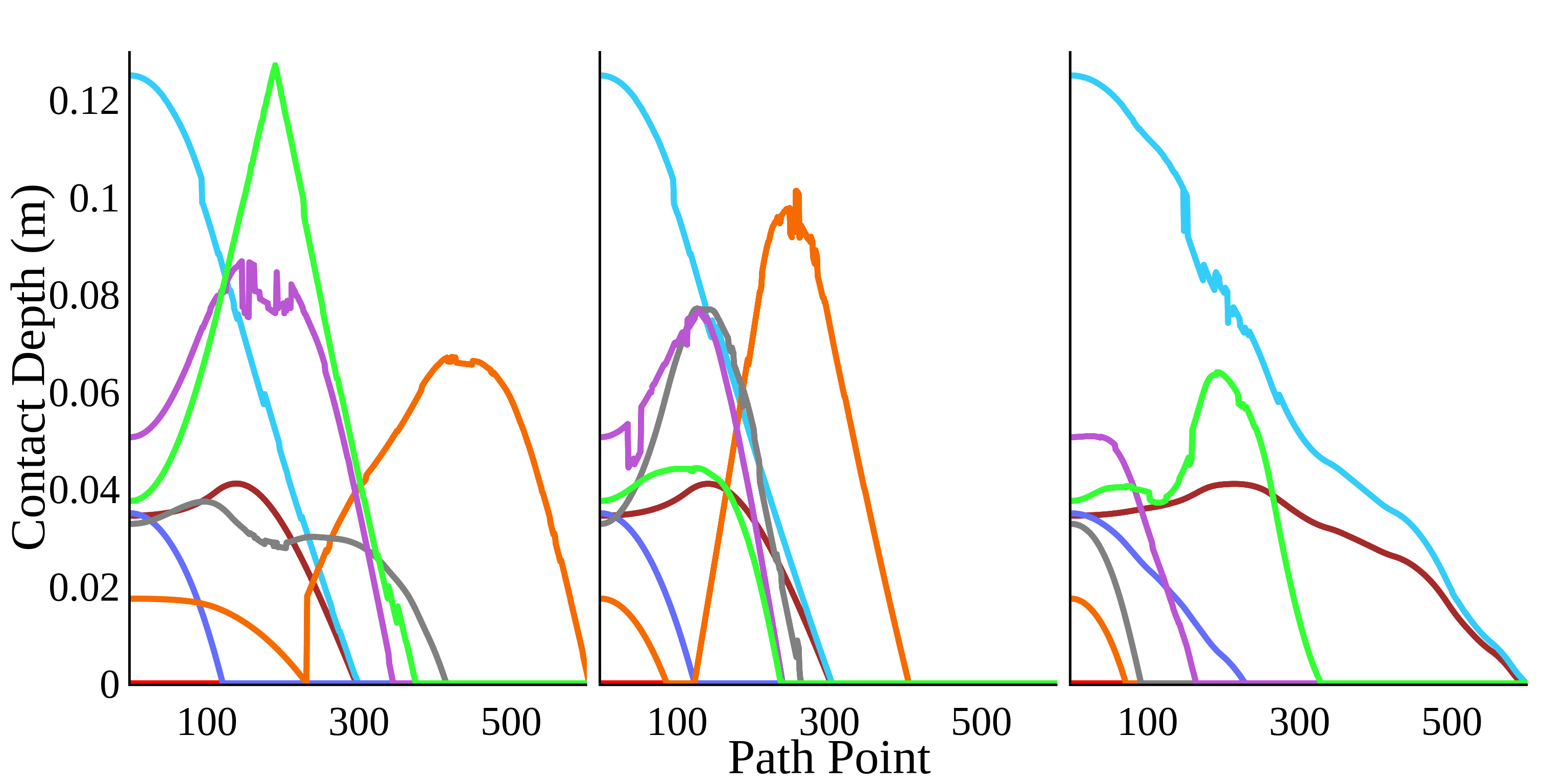}%
    \put(22, 40){\large\color{black}RRT*}
    \put(50, 40){\large\color{black}BIT*}
    \put(78, 40){\large\color{black}CAT-RRT}
    \put(45, 48)
    {\includegraphics[width=0.35\linewidth]{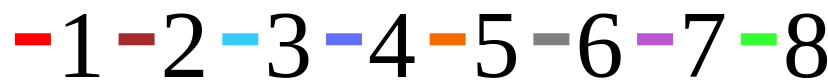}}
    \put(34, 48){\large\color{black}Link}
    \end{overpic}\vspace{-4pt}
        \caption{Contact depth at each link along one generated sample trajectory in Scenario 4. The peaks correspond to a high level of overlap between the link and the obstacle. Whereas RRT* and BIT* have three and two peaks each, CAT-RRT maintains one prolonged contact at a single link, achieving a faster convergence to goal with a shorter path length.}
        \label{fig:scene4-results}
\end{figure}

The results from Scenario 4 are also detailed on \cref{T:scene34-results}. In this scenario, CAT-RRT outperforms the other planners in its ability to generate a path of shortest length, with the least impact, and faster computationally. On \cref{fig:scene4-results}, the contact penetration depth at every link is plotted across a sample trajectory generated by each planner. With only one peak, as opposed to two and three for BIT* and RRT* respectively, CAT-RRT demonstrates its ability to keep one link in contact while moving other links through free space. This concept is highlighted in \cref{fig:first-fig}. This is another reason for which the path length of CAT-RRT is shorter, as it can maintain contact with the obstacle at the base while moving perpendicular to the obstacle at the end-effector. In contrast, the other planners tend to produce contact more randomly along the links while searching for a minimum-cost path to the goal. This results in longer high-cost trajectories for the arm and each link.  

\subsection{Real-world demonstration} \label{subsec:real-world}
To validate our simulation findings, we set up a real-world experiment that demonstrates a situation in which contact is harmless. More specifically, we show how the robot can reach for an object while making contact with a soft obstacle which overlaps with the goal state. Our supplementary video showcases the results \cite{projectlink}. Although the planning time of CAT-RRT remains a challenge for real-world operation, we believe this can be addressed with parallel computing and algorithm optimization.  

\section{Conclusion and Future Work} \label{conclusion}
This work is guided by the idea that planners can enhance their operational capabilities by reducing reliance on collision checking and increasing tolerance to contact with objects. We present a method that allows robots to intelligently plan for contact given a limited understanding of the environment. We show that our planner can successfully generate a path into high-cost regions with obstacles. Compared to other planners, which use a single cost for the entire arm, CAT-RRT can create shorter paths in less time using a per-link cost heuristic.
In our future research, we aim to demonstrate how robots can help leverage more “action” in the “sense-perceive-act” paradigm \cite{bohg2017interactive}. To do so, we aim to tightly couple CAT-RRT with control (\cite{escobedo2021contact,escobedo2022object}) to track contact during trajectory execution and perception to adjust planning costs based on object properties (e.g. hard or soft material). 
We believe that a robot that can plan and adjust for contact is better equipped to handle manipulation tasks in unstructured environments in the agricultural, industrial, and retail sectors. 
\AtNextBibliography{\small}
\printbibliography
\end{document}